\documentclass[conference]{IEEEtran}
\usepackage{times}

\usepackage[numbers]{natbib}
\usepackage{multicol}
\usepackage[bookmarks=true]{hyperref}
\hypersetup{
    colorlinks=true,
    linkcolor=blue,
    filecolor=magenta,      
    urlcolor=cyan,
    }
\usepackage{amssymb}
\usepackage{amsmath}
\usepackage{caption}
\usepackage{subcaption}
\usepackage{graphicx}

\newcommand{\obs}{x}
\newcommand{\act}{y}

\newcommand{\actsamples}{\{\tilde{\act}_i^m\}_{m=1}^M}
\newcommand{\Expect}[2][{p_\theta(\act|\obs_i)}]{{\mathbb{E}_{#1}\left[{#2}\right]}}
\newcommand{\grad}[2][\theta]{{\nabla_{#1}{#2}}}
\newcommand{\Ebm}{{E_\theta(\obs, \act)}}
\newcommand{\EbmXiYi}[1][\theta]{{E_{#1}(\obs_i, \act_i)}}
\newcommand{\EbmXiY}[1][\theta]{{E_{#1}(\obs_i, \act)}}
\newcommand{\EbmXiYm}[1][\theta]{{E_{#1}(\obs_i, \tilde{\act}_i^m)}}
\newcommand{\mean}[2][i]{{\frac{1}{#2}\sum_{#1=1}^{#2}}}
\newcommand{\prob}{{p_\theta(\act|\obs)}}
\newcommand{\probi}{{p_\theta(\act|\obs_i)}}

\newcommand{\logprob}{{\log p_\theta(\act|\obs)}}
\newcommand{\logprobi}{{\log p_\theta(\act|\obs_i)}}
\newcommand{\logprobii}{{\log p_\theta(\act_i|\obs_i)}}

\newcommand{\logZxi}{{\log Z_\theta(\act_i)}}
\newcommand{\entropy}{\mathcal{H}}
\newcommand{\entropyprob}{{\entropy(p_\theta(\act|\obs)})}

\DeclareMathOperator*{\argmin}{arg\,min}

\newcommand{\footref}[1]{\textsuperscript{\ref{#1}}}

\begin{document}

\title{Conditional Energy-Based Models for Implicit Policies: The Gap between Theory and Practice}

\author{Duy-Nguyen Ta, Eric Cousineau, Huihua Zhao, and Siyuan Feng \\
Toyota Research Institute \\
Cambridge, Massachusetts, USA \\
Email: eric.cousineau@tri.global}


\maketitle

\begin{abstract}
We present our findings in the gap between theory and practice of using conditional energy-based models (EBM) as an implicit representation for behavior-cloned policies. We also clarify several subtle, and potentially confusing, details in previous work in an attempt to help future research in this area. We point out key differences between unconditional and conditional EBMs, and warn that blindly applying training methods for one to the other could lead to undesirable results that do not generalize well. Finally, we emphasize the importance of the Maximum Mutual Information principle as a necessary condition to achieve good generalization in conditional EBMs as implicit models for regression tasks.
\end{abstract}

\IEEEpeerreviewmaketitle

\section{Introduction}

With many intriguing properties recently shown in \cite{florence2022implicit}, conditional energy-based models (EBMs) have garnered significant interest in the robotics manipulation community as an implicit representation for behavior-cloned policies to deal with multimodal and inconsistent demonstrations. Unfortunately, training conditional EBMs for regression problems can be challenging in practice, despite lots of remarkable successes of its "cousin" approach, unconditional EBMs, which are typically used as generative models for data on very high dimensional and nonlinear manifolds, e.g. images \cite{Ho2020nips,du2019implicit,dhariwal2021diffusion}. 

This paper does not present any new methods, and provides only minor empirical tuning on top of previous work. Rather, we aim to clarify the gap between theory and practice in training conditional EBMs for regression problems, such as behavior cloning. Our path to improving conditional EBM training was filled with many dead-ends due to incorrect and confusing exposition in the previous text \cite{florence2022implicit,gustafsson2020train} and our failure to realize the differences between conditional and unconditional EBMs, leading to unsuccessful attempts to apply methods for training unconditional EBMs to the other. We hope our lessons shared in this paper help clear the confusion, and prevents other people from making the same mistakes, and steer future research back to the right direction.

\section{Background}

Instead of representing a policy as an explicit function $\act=F_\theta(\obs)$, directly mapping from an observation $\obs$ to an action $\act$, Implicit Behavior Cloning (IBC) \cite{florence2022implicit} proposes to first learn an energy function $\Ebm$ that maps an observation-action pair to an energy value in $\mathbb{R}$. For inference, this energy function is minimized to produce a (hopefully) optimal action: $\hat{\act} = \argmin_{\act} \Ebm $. The intriguing benefits of this policy representation scheme have been discussed in \cite{florence2022implicit}.

In IBC, training the policy entails learning the energy function $\Ebm$ from a dataset of observation-action pairs collected from human demonstrations. As presented in \cite{florence2022implicit}, the loss function for training $\Ebm$ is the negative log likelihood (NLL): $\mathcal{L}_{NLL} = \sum_{i=1}^N-\logprobii$, in which the probability density function $\logprob$ relates to the EBM function $\Ebm$ via the Gibbs distribution: $\prob=\exp({-\Ebm})/Z_\theta(\obs)$, where $Z_\theta(\obs) = \int\exp(-\Ebm)d\act$ is the normalization factor depending on the given $\obs$. In short,
\begin{equation}
\mathcal{L}_{NLL}(\theta) = \sum_{i=1}^N-\log\frac{\exp(-\EbmXiYi)}{\int\exp(-\EbmXiY) d\act}. \label{eq:NLL-loss}
\end{equation}

NLL loss is also very commonly used in most unconditional EBM works \cite{lecun2006tutorial,song2021train}, whose main concerns are different ways to deal with the intractable integral in the normalization factor. In \cite{florence2022implicit}, the nuisance integral is approximated as 
\begin{equation}
\int e^{-\EbmXiY} d\act \approx e^{-\EbmXiYi} + \sum_{m=1}^M e^{-\EbmXiYm},\label{eq:int-approx}
\end{equation}
where $\actsamples$ is a set of negative action samples for each observation $\obs_i$. This leads to the following loss function:
\begin{equation}
\mathcal{L}_{InfoNCE}(\theta)= \sum_{i=1}^{N}-\log\frac{e^{-\EbmXiYi}}{e^{-\EbmXiYi} + \sum_{m=1}^M e^{-\EbmXiYm}}.\label{eq:InfoNCE}
\end{equation}

Different sampling schemes could be used to obtain negative samples $\actsamples$ from the action domain. Beside the uniform sampling scheme in Derivative-Free Optimizer (DFO) method, \cite{florence2022implicit} also proposes to use Langevin MCMC to sample directly from the currently learned distribution $\probi$.

\section{Common Misinterpretations}

The InfoNCE loss (\ref{eq:InfoNCE}) actually has a different meaning, which we will revisit in Section \ref{sec:MaxMI}. Here, we discuss the problems of interpreting it as an approximation of the NLL loss as currently presented. In fact, the approximation of the integral using a set of negative samples in (\ref{eq:int-approx}) raises many questions. First, a set of samples is typically used to approximate the expectation of an arbitrary function: $\Expect[p(y)]{f(y)} = \int f(y) p(y) dy \approx \frac{1}{M} \sum_m f(y^m)$, where $y^m \sim p(y)$. However, the normalization factor is \textbf{not} an expectation but the integral of the function itself, hence the formula does not directly apply. A popular technique to better approximate the integral of a function $\int f(y) dy$ is to use importance sampling \cite{gustafsson2020train}, which turns the integral into an expectation under a chosen proposal distribution $q(y)$: $ \int f(y) dy = \int \frac{f(y)}{q(y)}q(y) dy \approx \frac{1}{M} \sum_m \frac{f(y^m)}{q(y^m)} $, where $y^m \sim q(y)$, and the weight $w_m=1/q(y^m)$ of each sample have to be taken into account. Because of this, the approximation in (\ref{eq:int-approx}) is only correct if the proposal $q(y)$ is a uniform distribution, as in DFO. Unfortunately, as noted in prior work, uniform sampling does not scale well for problems with high dimensional action spaces \cite{florence2022implicit}.\footnote{We tried using Gaussian mixture models as the proposal density, following \cite{gustafsson2020train}, but did not achieve good results for problems in high dimensional spaces.}

We find that leveraging Langevin MCMC to obtain negative samples from the currently learned distribution $\probi$ to approximate the integral in (\ref{eq:int-approx}) is also confusing in several other ways, not just because the sample weights are not taken into account. The current set of samples approximating the current distribution becomes outdated when the distribution is updated in the next optimization step. To be correct, experiments (in addition to what is indicated in the IBC code) should be performed doing backpropagation through the entire chain of MCMC; however, this is a challenging task due to the stochasticity and discrete nature of the samples. A similar problem involving optimization of an integral of the distribution under optimization also arises in Variational Auto Encoder, where the reparameterization trick is used to obtain samples from a (different) \textit{fixed} normal distribution \cite{kingma2019introduction}. Unfortunately, the technique is not applicable in our case, since the inverse distribution required in the reparameterization trick does not have a closed form formula in our general setting.

In EBM literature, samples of the currently learned distribution from Langevin MCMC are used to approximate the gradient of the NLL loss, not the integral in the loss itself. This effectively leads to another loss function whose gradient w.r.t. $\theta$ is the same as the gradient of (\ref{eq:NLL-loss}) approximated using the set of MCMC samples \footnote{Detailed derivation can be found in \cite{lecun2006tutorial}}:
\begin{equation}
\mathcal{L}_{MCMC}(\theta)=\sum_{i=1}^{N}\left(E_{\theta}(x_{i},y_{i})-\frac{1}{M}\sum_{m=0}^{M}E_{\theta}(x_{i},\tilde{y}_{i}^{m})\right).\label{eq:MCMC-loss}
\end{equation}
Unfortunately, we did not notice significant improvements in using this loss function in our experiments.

\begin{figure}
     \centering
     \begin{subfigure}[b]{0.2\textwidth}
         \centering
         \includegraphics[width=\textwidth]{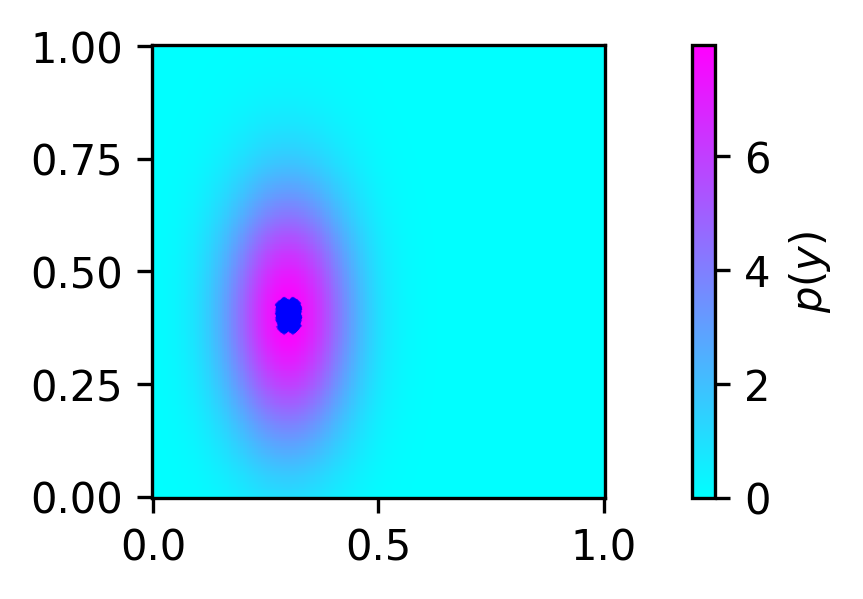}
         \caption{IBC's Langevin samples}
         \label{fig:distr_langevin_ibc}
     \end{subfigure}
     \begin{subfigure}[b]{0.2\textwidth}
         \centering
         \includegraphics[width=\textwidth]{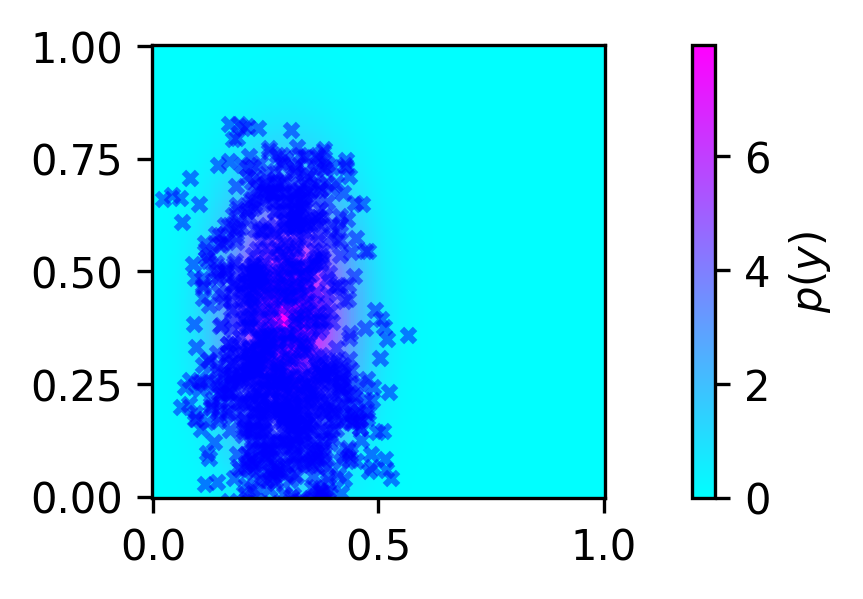}
         \caption{True Langevin samples}
         \label{fig:distr_langevin_true}
     \end{subfigure}
     \caption{IBC's vs True Langevin MCMC samples\footref{footnote:langevin}}
     \label{fig:distr_langevin_compare}
     \vspace{-18pt}
\end{figure}

Another important mismatch in theory and practice lies in the Langevin MCMC formula used in IBC. The correct formulation \cite{du2019implicit,girolami2011riemann,wiki2022mala} is
\begin{gather}
y^{k+1} = y^k - \frac{\lambda}{2} \nabla_y E_{\theta}(x,y^k) + \sqrt{\lambda}\ \omega^k, \hspace{10pt}
\omega^k \sim \mathcal{N}(0, I) \label{eq:langevin-correct}
\end{gather}
However, in IBC, the following formula is effectively used:
\begin{gather}
y^{k+1} = y^k - \frac{\lambda}{2} \nabla_y E_{\theta}(x,y^k) + \lambda\ \hat{\omega}^k,
\hspace{10pt}
\hat{\omega}^k \sim \mathcal{N}(0, \sigma) \label{eq:langevin-ibc}
\end{gather}
where $\sigma$ permits additional scaling of the noise term.
Even assuming $\sigma = 1$, the subtle difference between the coefficients for gradient and noise terms can create a large difference in how the samples approximate the underlying distribution. Fig. \ref{fig:distr_langevin_compare} shows that the correct formulation approximates the mean and variance of the Gaussian accurately, whereas the IBC formulation does not.
\footnote{Details on the correct and incorrect Langevin formulation, numerical experiments, and plots can be found in Appendix \ref{appendix:langevin-formulation}. \label{footnote:langevin}}

However, in training EBM, we found that the IBC's incorrect Langevin MCMC formulation seems to work \textit{better} in practice than the correct formulation. For some of the tasks we tested against, we found it useful to tune the noise scaling $\sigma$ of IBC's Langevin MCMC down to a small value, typically 0.01, so that the negative samples are very close to the positive samples for good training results. We also found that using one long chain MCMC, throwing away initial samples during the burn-in stage, and collecting $M$ samples at the end of the chain, as typically done in MCMC literature, does not work as well as using multiple independent short chains and selecting the last sample from each chain, as is done in IBC.

These mismatches between theory and practice reveal the gap in our understanding of Langevin MCMC in conditional EBMs training, and we await a better theoretical explanation. 

\section{Failed Attempts}\label{sec:failed_attempts}

To improve EBM's robustness and generalization, we first tried to avoid a well-known ``deep valley'' problem, where the low energy region around a positive sample is very small and closely surrounded by the high energy regions around it \cite{lecun2006tutorial}. In fact, the ideal Gibbs distribution that minimizes the NLL loss in (\ref{eq:NLL-loss}) is a mixture of Dirac delta distributions whose peaks are at the positive data points and volume under the curve on the entire domain (i.e. the integral in the denominator of (\ref{eq:NLL-loss})) is zero. Such a function, however, might suffer from robustness and generalization issues. Following MaxEnt RL \cite{eysenbach2021maximum}, we added a maximum conditional entropy regularizer to the original loss function, detailed in the Appendix, which minimizes the variance of energies of negative samples, effectively forcing the resulting distribution to be close to a uniform distribution, which has the maximum entropy. We observe that the correct Langevin MCMC can produce reasonable results using this regularizer, however, it is not better than the original IBC's Langevin MCMC with a small noise $\sigma$.

We also tried to improve the quality of MCMC samples by using a replay buffer to simulate long-chain MCMCs \cite{du2019implicit}. This technique for training unconditional EBMs does not directly apply to conditional EBMs, since different conditional distributions cannot share the same replay buffer. Hence, we train a joint EBM, modeling the joint distribution $p_\theta(\obs,\act)$, which is also reducible to the conditional for inferring the optimal action $p_\theta(\obs_i,\act)\propto \probi$. The results show that it can overfit to the training data very well only after a few epochs, but generalizes very poorly to unseen data in the validation set. This should not come as a surprise because all samples of unseen observations in the joint domain are treated as negative samples during training. The lessons we learned from this failure are two-fold. First, blindly applying techniques for unconditional EBMs to train conditional EBMs might not be a good idea. Second, there are many models with the same representational power that can fit the training data well, but not all of them can generalize well to unseen observations.
Although the IBC's incorrect Langevin MCMC happens to generalize well, a principle to select implicit models with better generalization capability is currently lacking.
 
\section{Maximum Mutual Information}\label{sec:MaxMI}

We realize the following key difference between unconditional and conditional EBMs in their requirements for generalization: whereas unconditional EBMs for generative model are meant to sample new data from a learned distribution, conditional EBMs for regression requires the inference process to produce outputs that generalize well to unseen inputs. 

A necessary condition for good generalization in regression is that the model should maximize the dependencies between the input and output. Hence, one of the goals of the loss function for training conditional EBMs should be to maximize the mutual information between them to improve generalization. In fact, the InfoNCE loss function in (\ref{eq:InfoNCE}) was originally derived for that purpose \cite{van2018representation}. It is \textbf{not} an approximation of the NLL loss, but the negative of a lower bound of the mutual information between observations and actions. Minimizing it effectively maximizes that lower bound, resulting in distributions with high mutual information. Due to that effect, it underlies many successful results in self-supervised contrastive learning, e.g. \cite{chen2020simple,radford2021learning}.

However, the InfoNCE loss function has to be used with care. Essentially, it models the ratio between the conditional and the marginal densities in the mutual information $I(\obs,\act)=\int p(\obs,\act)\log\frac{\prob}{p(\act)}d\obs d\act$. Part of its name ``noise contrastive estimation'' (NCE) stems from the goal to differentiate noisy samples of $p(\act)$ from the true samples of $\prob$. Following the rigorous proof in \cite{poole2019variational}, \cite{tschannen2019mutual} pointed out that in order for InfoNCE to be a proper bound of the mutual information, the negative samples $\actsamples$ have to be independently sampled from the marginal $p(\act)$. This explains why our attempts to obtain the correct samples of the currently learned conditional distribution using a long-chain MCMC did not go well: they are not independent and do not come from a noise distribution. Using independent samples from multiple short chains as done in IBC has a better chance to succeed.

Moreover, the small noise scaling $\sigma$ in IBC's incorrect Langevin MCMC actually helps in maximizing the mutual information. Due to the small $\sigma$, many generated negative samples are very close to the positive ones, leading to the aforementioned ``deep valley'' effect. However, despite being undesirable in unconditional EBMs, these deep valleys should be favorable in conditional EBMs since they produce conditional densities with very low entropy $\entropy(\act|\obs)$, effectively increasing the mutual information, since $I(\obs,\act)=\entropy(\act)-\entropy(\act|\obs)$. Explicit models are the extreme case where $\prob$ are Dirac delta distributions with zero entropy. This explains why our earlier attempts to maximize the conditional entropy following MaxEnt RL literature headed to a wrong direction.

\begin{figure}
    \centering
    \includegraphics[width=0.45\textwidth]{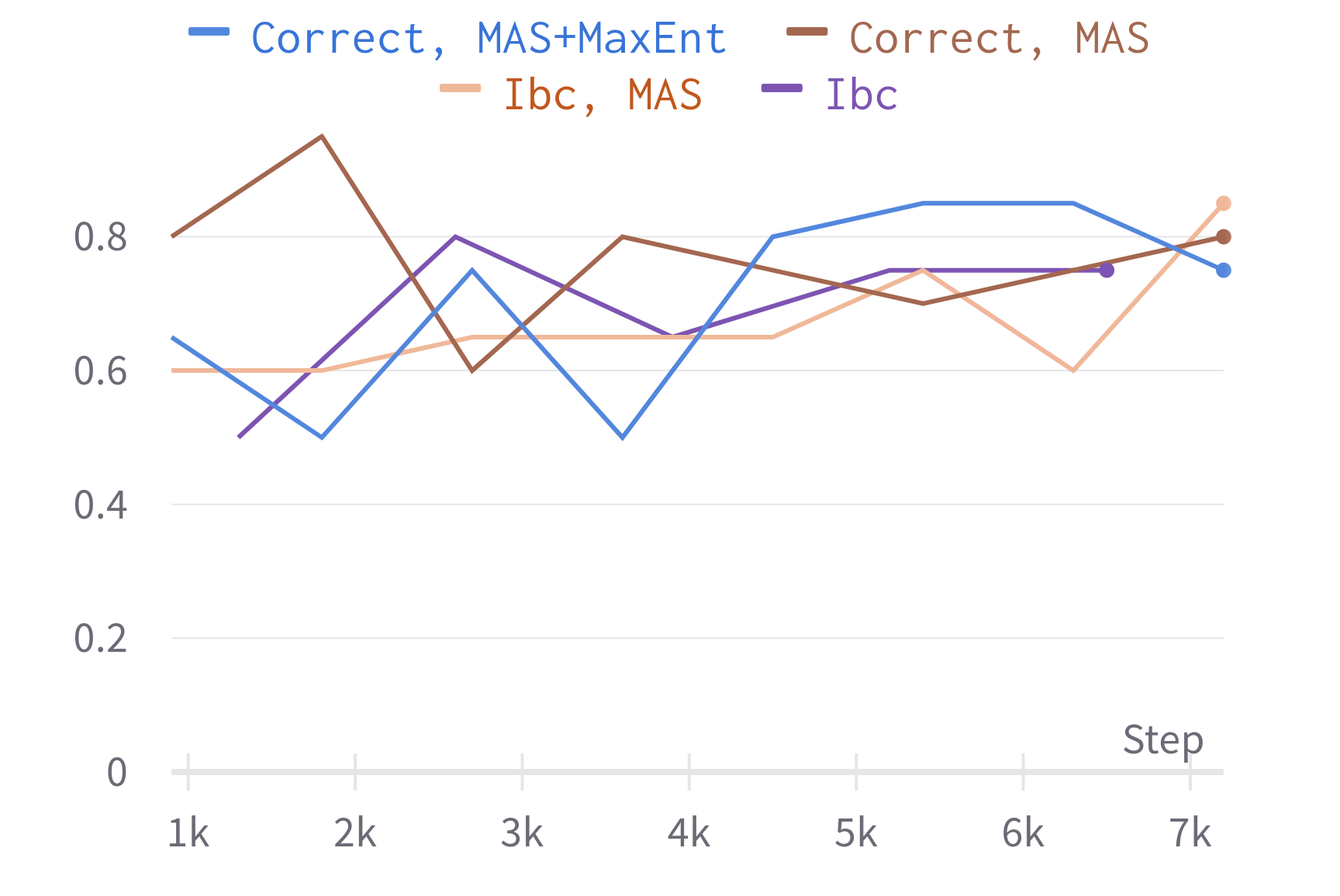}
    \caption{Validation success rates of IBC's Langevin (cyan) and Marginal Action Sampler (red) \footref{footnote:results}}
    \label{fig:train_ibc_vs_mas}
     \vspace{-18pt}
\end{figure}

Inspired by the mutual information lower-bound proof of InfoNCE, we also tried to replace the IBC's incorrect Langevin sampler with a marginal action sampler (MAS) to obtain independent negative samples from $p(\act)$. We train the marginal EBM $p_\gamma(\act)$ using the correct Langevin MCMC method with the maximum entropy regularization (see the Appendix). Maximizing the marginal entropy $\entropy(\act)$ helps to further increase the mutual information (see the $I(\obs,\act)$ formula above). We use MCMC samples from training the marginal action EBM $p_\gamma(\act)$ as negative samples in the InfoNCE loss (\ref{eq:InfoNCE}) to train the targeted conditional EBM concurrently. Our preliminary results on particle environments in \cite{florence2022implicit} are encouraging. As shown in Fig. \ref{fig:train_ibc_vs_mas}, EBMs trained by MAS appear to be on par with IBC's incorrect Langevin MCMC. \footnote{See Appendix \ref{appendix:training-results} for more details about the training setup. \label{footnote:results}}

\section{Conclusion and Future work} 
\label{sec:conclusion}

We identified the gap between theory and practice in training conditional EBMs for implicit policy representation, and clarified the confusing interpretation in previous work. We also pointed out the key difference in generalization requirements between unconditional and conditional EBMs, which calls for more attention on conditional EBMs training methods. We arrived at the Maximum Mutual Information principle to improve the generalization of conditional EBMs for regression. However, that principle might not be sufficient. In order for the model to generalize well, it should also capture the ``trend'' of how the input-output data in their joint domain spread far beyond the training data regions. Leveraging the generalization power of explicit functions to improve that of the implicit ones, e.g. \cite{xie2021cooperative}, is an interesting direction for future work.

\section*{Acknowledgments}

Pete Florence for valuable discussions, and all Google IBC authors for their excellent work.

\bibliographystyle{plainnat}
\bibliography{references}

\clearpage

\onecolumn

\appendix
\section{Appendix}

\subsection{Derivation of Maximum Entropy Regularization}

Here we derive the maximum entropy ($\entropy$) regularizer in the negative log-likelihood loss function, used in our failed attempt in Section \ref{sec:failed_attempts} and in the Marginal Action Sampler in Section \ref{sec:MaxMI}:
\begin{align*}
\mathbb{-E}_{p_{data}}\left[\log(p_{\theta}(y|x))+\entropy(p_{\theta}(y|x))\right] & \approx-\frac{1}{N}\sum_{i=1}^{N}\left(\log(p_{\theta}(y_{i}|x_{i}))-\int p_{\theta}(y|x_{i})\log(p_{\theta}(y|x_{i}))dy\right)\\
 & =-\frac{1}{N}\sum_{i=1}^{N}\left(\log\left(\frac{\exp(-E_{\theta}(x_{i},y_{i}))}{Z_{\theta}(x_{i})}\right)-\int p_{\theta}(y|x_{i})\log\left(\frac{\exp(-E_{\theta}(x_{i},y))}{Z_{\theta}(x_{i})}\right)dy\right)\\
 & =-\frac{1}{N}\sum_{i=1}^{N}\left(-E_{\theta}(x_{i},y_{i})-\log Z_{\theta}(x_{i})-\int p_{\theta}(y|x_{i})\left(-E_{\theta}(x_{i},y)-\log Z_{\theta}(x_{i})\right)dy\right)\\
 & =-\frac{1}{N}\sum_{i=1}^{N}\left(-E_{\theta}(x_{i},y_{i})-\log Z_{\theta}(x_{i})+\int p_{\theta}(y|x_{i})E_{\theta}(x_{i},y)dy+\int p_{\theta}(y|x_{i})\log Z_{\theta}(x_{i})dy\right)\\
 & =-\frac{1}{N}\sum_{i=1}^{N}\left(-E_{\theta}(x_{i},y_{i})-\log Z_{\theta}(x_{i})+\int p_{\theta}(y|x_{i})E_{\theta}(x_{i},y)dy+\log Z_{\theta}(x_{i})\right)\\
 & =\frac{1}{N}\sum_{i=1}^{N}\left(E_{\theta}(x_{i},y_{i})-\int p_{\theta}(y|x_{i})E_{\theta}(x_{i},y)dy\right),
\end{align*}

Taking the gradient of the loss:
\[
-\grad{\Expect[p_{data}]{\logprob + \entropyprob}} \approx \mean{N} \left( \grad{\EbmXiYi} -\grad{\int \probi \EbmXiY dy} \right).
\]

Expanding the second term:
\begin{align*}
    -\grad{\int\probi \EbmXiY dy} & =-\int \left( ( \grad{\probi} ) \EbmXiY + \probi \grad{\EbmXiY} \right) dy \\
    & =-\int (\grad{\probi}) \EbmXiY dy - \Expect{\grad{\EbmXiY}}  \\
    & =-\int \probi (\grad{\logprobi}) \EbmXiY dy - \Expect{\grad{\EbmXiY}}  \\
    & =-\Expect{(\grad{\logprobi}) \EbmXiY} - \Expect{\grad{\EbmXiY}}  
 \end{align*}
 
then further expanding the first term:
\begin{align*}
-\Expect{\EbmXiY \grad{\logprobi}} & = -\Expect{\EbmXiY \grad{(-\EbmXiY-\logZxi)}}
 \\
 & = \Expect{\EbmXiY \grad{\EbmXiY} + \EbmXiY \grad{\logZxi}}
 \\
 & = \Expect{\EbmXiY \grad{\EbmXiY} } + \Expect{\EbmXiY} \grad{\logZxi}
 \\
 & = \Expect{\EbmXiY \grad{\EbmXiY} } - \Expect{\EbmXiY} \Expect{\grad{\EbmXiY}},
\end{align*}

where we have used the well-known fact that
\begin{align}
\nabla_{\theta}\log\int\exp\left(-E_{\theta}(x_{i},y)\right)dy & =\frac{1}{\int\exp\left(-E_{\theta}(x_{i},y)\right)dy}\nabla_{\theta}\int\exp\left(-E_{\theta}(x_{i},y)\right)dy\nonumber \\
 & =\frac{1}{Z_{\theta}(x_{i})}\int\nabla_{\theta}\exp\left(-E_{\theta}(x_{i},y)\right)dy\nonumber \\
 & =\int\frac{\exp\left(-E_{\theta}(x_{i},y)\right)}{Z_{\theta}(x_{i})}\nabla_{\theta}\left(-E_{\theta}(x_{i},y)\right)dy\nonumber \\
 & =\int p_{\theta}(y|x_{i})\nabla_{\theta}\left(-E_{\theta}(x_{i},y)\right)dy\nonumber \\
 & =\mathbb{E}_{p_{\theta}(y|x_{i})}\left[\nabla_{\theta}\left(-E_{\theta}(x_{i},y)\right)\right]\label{eq:expectation_of_grad}
\end{align}

In summary,
\begin{align*}
-\grad{\Expect[p_{data}]{\logprob + \entropyprob}} \approx \mean{N} & \left( 
    \grad{\EbmXiYi} - \Expect{\grad{\EbmXiY}} \right. 
\\
    &\left. \quad + \Expect{ \EbmXiY \grad{\EbmXiY} } \right.
\\
    &\left. \quad - \Expect{\EbmXiY} \; \Expect{\grad{\EbmXiY}}
     \right) 
\end{align*}

Monte Carlo approximation of the gradient:
\begin{align*}
-\grad{\Expect[p_{data}]{\logprob + \entropyprob}} \approx \mean{N} & \left( 
    \grad{\EbmXiYi} -  \mean[m]{M}\grad{\EbmXiYm} \right. 
\\
    &\left. \quad +  \mean[m]{M}\EbmXiYm \grad{\EbmXiYm} \right.
\\
    &\left. \quad -  \mean[m]{M}\EbmXiYm \;  \mean[m]{M}\grad{\EbmXiYm}
     \right) 
\end{align*}
where $\tilde{y}_{i}^{m}\sim p_{\theta}(y|x_{i})$.

The new loss function that has the same gradient is:

\begin{align*}
\mathcal{L}_{entropy}(\theta) & = \mean{N} \left( \EbmXiYi - \mean[m]{M}\EbmXiYm + \frac{1}{2}\mean[m]{M}\left(\EbmXiYm\right)^2 - \frac{1}{2}\left( \mean[m]{M}\EbmXiYm \right)^2 \right)
\end{align*}

Intuitively, it is similar to the traditional Langevin MCMC loss function \eqref{eq:MCMC-loss} with the added two terms that equal to the variance of the negative samples' energy values ($\text{Var}[E_\theta] = \mathbb{E}[E_\theta^2] - \mathbb{E}[E_\theta]^2)$. Minimizing the energy function's variance effectively makes the resulting EBM close to the uniform distribution.

\subsection{Langevin Correctness}
\label{appendix:langevin-formulation}

There seemed to have been a small transcription error when the authors were implementing the Langevin step in both math and code. To illustrate, we compare three slightly different formulations existing, with minor adjustments to notation for consistency and equating $E_{\theta}(y^k) = E_{\theta}(x,y^k)$. The step sizes used in each are either $\epsilon$, $\lambda$, or $\tau$ (where $\lambda = \epsilon^2 = 2 \tau$), and are ultimately equivalent:
\begin{enumerate}
    \item Equation 1 in \cite{du2019implicit} effectively states:
        \begin{align*}
        y^{k+1}
            &= y^k - \frac{\lambda}{2} \nabla_y E_{\theta}(y^k) + \psi^k,
                & \psi^k \sim \mathcal{N}(0, \lambda) \nonumber \\
            &= y^k - \frac{\lambda}{2} \nabla_y E_{\theta}(y^k) + \sqrt{\lambda}\ \omega^k,
                & \omega^k \sim \mathcal{N}(0, I)
        \end{align*}
    \item Equation 2 in \cite{girolami2011riemann} effectively states:
        \begin{align*}
        y^{k+1}
            &= y^k - \frac{\epsilon^2}{2} \nabla_y E_{\theta}(y^k) + \epsilon\ \omega^k,
                & \omega^k \sim \mathcal{N}(0, I)
        \end{align*}
    \item The Euler-Maruyama update rule from \cite{wiki2022mala} effectively states (with a sign-flip):
        \begin{align}
        y^{k+1}
            &= y^k - \tau \nabla_y E_{\theta}(y^k) + \sqrt{2 \tau}\ \omega^k,
                & \omega^k \sim \mathcal{N}(0, I) \label{eq:langevin-correct-impl}
        \end{align}
\end{enumerate}

However, while holding $\sigma = 1$, the equation in Sec. B.3 of \cite{florence2022implicit} effectively states:
\begin{align*}
y^{k+1}
    &= y^k - \frac{\lambda}{2} \nabla_y E_{\theta}(y^k) + \lambda\ \omega^k,
        & \omega^k \sim \mathcal{N}(0, I)
\end{align*}
We believe the transcription error happened where $\mathcal{N}(0, \lambda^2) = \lambda\ \mathcal{N}(0, I)$ was mistakenly equated with $\mathcal{N}(0, \lambda) = \sqrt{\lambda}\ \mathcal{N}(0, I)$.

Anecdotally, we believe that IBC's incorrect Langevin MCMC is closer to gradient descent than true Langevin MCMC, especially when the step size $\lambda$ is "far" from 1. Because the step sizes tend to be smaller, due to the normalization IBC employs, then the gradient term dominates and the noise term is diminished.

The plots in Fig. \ref{fig:distr_langevin_compare} show the final 1000 iterations from a total of 4000 iterations for a single chain. The code (and parameters) to generate the results can be found in this \href{https://github.com/EricCousineau-TRI/repro/blob/96886ea/python/torch/langevin_step.ipynb}{Jupyter notebook}.

\subsection{Implementation Details}

For inclusion in our codebase, we implemented Implicit Behavior Cloning in PyTorch, cross-referencing the published implementation in (which was using TensorFlow v2). We implemented both the DFO and Langevin based policies, and then built upon them for above experiments. We validated our implementation by comparing both overfit (1 episode) and "full" training (2000 episodes) for the 2d and 16d particle environment and ensured that the logged metrics (loss, energies, distances between positive and negative samples, success metrics, etc.) followed similar evolutions along training minibatches.

When we refer to experiments running correct Langevin, we actually use the formulation as stated in \eqref{eq:langevin-correct-impl} rather than \eqref{eq:langevin-correct}, meaning the step size indicated is actually $\tau = \frac{\lambda}{2}$. \footnote{We used \eqref{eq:langevin-correct} in the paper to provide clearer distinction against \eqref{eq:langevin-ibc}.}

\subsection{Training Setup}
\label{appendix:training-results}

Our primary demonstration environment is the 2d particle environment for its simplicity and speed of training.

For training, we used a very similar setup as those in \cite{florence2022implicit}:

\begin{itemize}
    \item We use the same normalization for the observation and action space, such that the minimum and maximum in each space occur at $[-1, 1]$, respectively.
    \item Energy network is a simple MLP, 256 hidden units, 2 layers\footnote{$(n_{in} \times n_{hidden}) \rightarrow (n_{hidden} \times n_{hidden}) \rightarrow (n_{hidden} \times n_{out})$}, 0\% dropout, ReLU activation, using Keras "normal"-style initializer\footnote{Linear weights and biases initialized from $\mathcal{N}(0, \sigma)$ for $\sigma = 0.05$}. For our experiments, we did not need the spectral norm application.
    \item For negative samples, we use Langevin sampling, either ``correct'' \eqref{eq:langevin-correct-impl} or ``ibc'' \eqref{eq:langevin-ibc}. For speed (both in training and inference), we use 10 iterations.
    \begin{itemize}
        \item For training, we select $\sigma = 0.1$ with ``ibc'' to allow more for aggressive convergence with fewer iterations.
        \item For inference, we select $\sigma = 0.01$ with ``ibc'' such that it behaves as gradient descent.
        \item For both training and inference, we use both "global" clipping and per-iteration clipping at 25\% of the original action space. When drawing initial samples from a uniform distribution, we apply the same 5\% margin to both ends of the action space as is done in \cite{florence2022implicit} (thus sampling from $[-1.1, 1.1]$).
        \item For the step size ($\lambda$ for ``ibc'', $\tau$ for ``correct''), we use the same polynomial scheduler, starting at step size of 1, decreasing to 0.001, with a power of 2.
        \item Due to a minor bug, we did not end up using gradient penalty as indicated in Sec. B.3.1 in \cite{florence2022implicit}.
    \end{itemize}
    \item Training used PyTorch's Adam optimizer with learning rate of 0.001, batch size of 512, step decay learning rate scheduler with gamma of 0.99 decaying every 100 epochs.
    \item The following distinctions are made for the following trials, with 300 episodes for training:
    \begin{itemize}
        \item \textbf{Ibc} - Running ``ibc`` Langevin formulation as mentioned above.
        \item \textbf{Ibc, MAS} - Using Marginal Action Sampler (same MLP architecture, but using 64 hidden units).
        \item \textbf{Correct, MAS} - Same as prior, but using ``correct'' Langevin formulation.
        \item \textbf{Correct, MAS+MaxEnt} - Same as prior, but using maximum entropy as well as an L2 norm on positive energies (anchoring them towards zero).
    \end{itemize}
\end{itemize}

\end{document}